\documentclass[conference]{IEEEtran}

\usepackage{cite}
\usepackage{amsmath,amssymb,amsfonts}
\usepackage{amsthm}
\usepackage{algorithmic}
\usepackage{graphicx}
\usepackage{textcomp}
\usepackage{xcolor}
\usepackage{subcaption}
\usepackage{multirow}
\usepackage{booktabs}
\newcommand{\ra}[1]{\renewcommand{\arraystretch}{#1}}
\def\BibTeX{{\rm B\kern-.05em{\sc i\kern-.025em b}\kern-.08em
    T\kern-.1667em\lower.7ex\hbox{E}\kern-.125emX}}

\begin{document}

\title{Using Cyber Terrain in Reinforcement Learning for Penetration Testing}

\author{Rohit Gangupantulu$^{a, \dagger}$,
        Tyler Cody$^{b, \dagger, *}$,
        Paul Park$^{a}$, 
        Abdul Rahman$^{c}$,\\
        Logan Eisenbeiser$^{b}$,
        Dan Radke$^{c}$, 
        Ryan Clark$^{c}$,
        Christopher Redino$^{c}$\\
        \small $^{a}$Deloitte Consulting LLC \\
        \small $^{b}$National Security Institute, Virginia Tech
        \small $^{c}$Deloitte \& Touche LLP \\
        \small $^{\dagger}$Co-First Authors \\
        \small $^{*}$Corresponding Author: Tyler Cody; tcody@vt.edu \\
}

\IEEEoverridecommandlockouts
\IEEEpubid{\makebox[\columnwidth]{978-1-6654-8356-8/22/\$31.00 \copyright 2022 IEEE \hfill}
\hspace{\columnsep}\makebox[\columnwidth]{ }}

\maketitle

\IEEEpubidadjcol

\begin{abstract}

Reinforcement learning (RL) has been applied to attack graphs for penetration testing, however, trained agents do not reflect reality because the attack graphs lack operational nuances typically captured within the intelligence preparation of the battlefield (IPB) that include notions of (cyber) terrain. In particular, current practice constructs attack graphs exclusively using the Common Vulnerability Scoring System (CVSS) and its components. We present methods for constructing attack graphs using notions from IPB on cyber terrain. We consider a motivating example where firewalls are treated as obstacles and represented in (1) the reward space and (2) the state dynamics. We show that terrain analysis can be used to bring realism to attack graphs for RL. We use an attack graph with roughly 1000 vertices and 2300 edges and deep Q reinforcement learning with experience replay to demonstrate the method. 

\end{abstract}


\section{Introduction}

Prediction of vulnerabilities and exploits to support cyber defense (i.e., the blue picture) typically involves analyzing ingested data  acquired from sensors and agents in various networks.  Learning behaviors based on what has been seen, i.e., observed on the network, involves meticulous curation and processing of this data to support model development, training, testing, and validation. While this has provided some degree of results in the past, this work intends to explore an alternative approach toward identifying weaknesses within networks. The goal is to leverage the attack graph construct \cite{mcdermott2001attack} and train machine learning models over them to predict weaknesses within network topologies.

Under this approach, instead of observing a static, curated data set, machine learning algorithms can learn by interacting with attack graphs directly. Reinforcement learning (RL) for penetration testing has shown this to be feasible given constraints on attack graph representation such as scale and observability. However, existing literature constructs attack graphs either with no vulnerability information \cite{ghanem2018reinforcement, schwartz2019autonomous, ghanem2020reinforcement, chaudhary2020automated} or entirely with vulnerability information \cite{yousefi2018reinforcement, chowdary2020autonomous, hu2020automated}.

Yousefi \textit{et al.}, Chowdary \textit{et al.}, and Hu \textit{et al.} use the Common Vulnerability Scoring System (CVSS) and its components to construct attack graphs \cite{yousefi2018reinforcement, chowdary2020autonomous, hu2020automated}, similar to Gallon and Bascou \cite{gallon2011using}. CVSS scores are an open, industry-standard means of scoring the severity of cybersecurity vulnerabilities. They provide an empirical and automatic means of constructing attack graphs for RL. However, they do not always correlate to a useful contextual picture for cyber operators. By relying totally on its abstractions, network representations unfortunately can be biased totally towards vulnerabilities and not on a realistic view of how an adversary plans or executes an attack campaign. As a result, this leads to RL methods converging to unrealistic attack campaigns.

While CVSS scores provide a strong foundation for attack graphs, we posit that notions of \emph{cyber terrain} \cite{conti_raymond_2018} should be built into attack graph representations to enable RL agents to construct more realistic attack campaigns during penetration testing. In particular, we suggest a focus on the OAKOC terrain analysis framework that consists of obstacles, avenues of approach, key terrain, observation and fields of fire, and cover and concealment \cite{conti_raymond_2018}. This work makes the following contributions:
\begin{itemize}
    \item We contribute methodology for building OAKOC cyber terrain into Markov decision process (MDP) models of attack graphs.
    \item We apply our methodology to RL for penetration testing by treating firewalls as cyber terrain obstacles in an example network that is at least an order of magnitude larger than the networks used by previous authors \cite{ghanem2018reinforcement, schwartz2019autonomous, ghanem2020reinforcement, chaudhary2020automated, yousefi2018reinforcement, chowdary2020autonomous, hu2020automated}.
\end{itemize}
In doing so we extend the literature on using CVSS scores to construct attack graphs and MDPs as well as the literature on RL for penetration testing.

The paper is structured as follows. First, background is given on terrain analysis and cyber terrain, reinforcement learning, and penetration testing. Second, our methods for constructing terrain-based attack graphs are presented. Then, results are presented before concluding with remarks on future steps.

\section{Background}

\subsection{Terrain Analysis and Cyber Terrain}

\begin{figure*}[t]
    \centering
    \includegraphics[width=\textwidth]{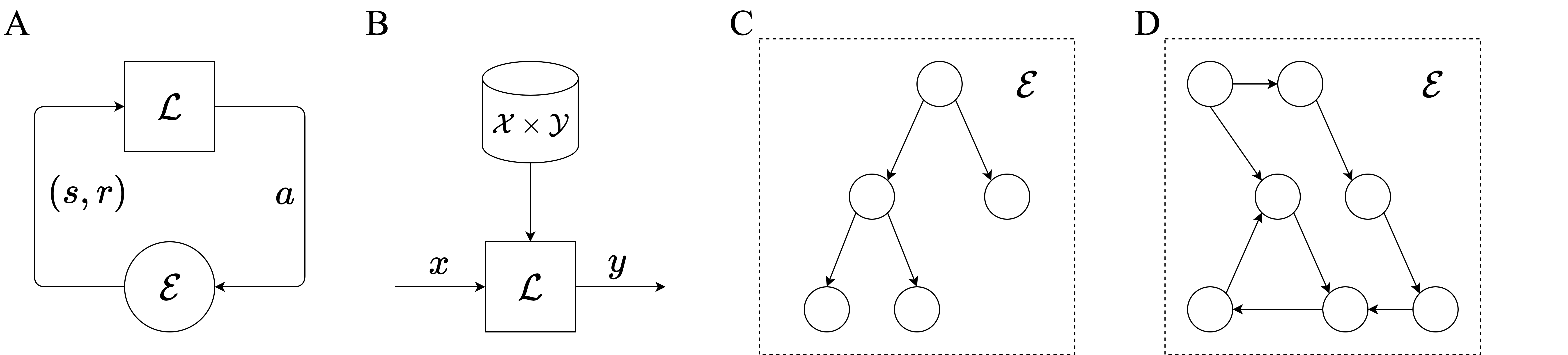}
    \caption{Figure \ref{fig:1}A shows a reinforcement learning agent $\mathcal{L}$ taking actions $a$ in environment $\mathcal{E}$ and receiving state $s$ and reward $r$. Figure \ref{fig:1}B shows a supervised learning agent $\mathcal{L}$ learning from example-label pairs $(x, y)$ provided by an oracle. Figure \ref{fig:1}C shows an environment $\mathcal{E}$ under the attack tree model. Figure \ref{fig:1}D shows an environment $\mathcal{E}$ under the attack graph model.}
    \label{fig:1}
\end{figure*}

Intelligence preparation of the battlefield (IPB) considers terrain a fundamental concept \cite{purcell1989operational}. In the physical domain, terrain refers to land and its features. Conti and Raymond define cyber terrain as, ``the systems, devices, protocols, data, software, processes, cyber personas, and other network entities that comprise, supervise, and control cyberspace \cite{conti_raymond_2018}.'' 

They note that cyber terrain exists at strategic, operational, and tactical levels, including, e.g., transatlantic cables and satellite constellations, telecommunications offices and regional data centers, and wireless spectrum and Local Area Network protocols, respectively. In this paper, we consider what Conti and Raymond refer to as the \emph{logical plane} of cyber terrain, which consists of data link, network, network transport, session, presentation, and application, i.e., layers 2-7 of the Open Systems Interconnection model \cite{zimmermann1980osi}.

Terrain analysis typically follows the OAKOC framework, consisting of observation and fields of fire (O), avenues of approach (A), key terrain (K), obstacles and movement (O), and cover and concealment (C). These notions from traditional terrain analysis can be applied to cyber terrain \cite{applegate2017searching}. For example, fields of fire may concern all that is network reachable (i.e., line of sight) and avenues of approach may consider network paths inclusive of available bandwidth \cite{conti_raymond_2018}. In this paper, we use obstacles to demonstrate how our methodology can be used to bring the first part of the OAKOC framework to attack graph construction for reinforcement learning.

\subsection{Reinforcement Learning}

Reinforcement learning is concerned with settings where agents learn from taking actions in and receiving rewards from an environment \cite{sutton2018reinforcement}. It can be contrasted with supervised learning, where agents learn from example-label pairs given by an oracle or labeling function. This contrast is depicted in Figures \ref{fig:1}A and \ref{fig:1}B. Naturally, reinforcement learning solution methods take a more dynamic formulation.

An agent is considered to interact with an environment $\mathcal{E}$ over a discrete number of time-steps by selecting an action $a_t$ at time-step $t$ from the set of actions $A$. In return, the environment $\mathcal{E}$ returns to the agent a new state $s_{t+1}$ and reward $r_{t+1}$. Thus, the interaction between the agent and environment $\mathcal{E}$ can be seen as a sequence $s_1, a_1, s_2, a_2, ..., a_{t-1}, s_t$. When the agent reaches a terminal state, the process stops.

Here we consider a case when $\mathcal{E}$ is a finite MDP. A finite MDP is a tuple $\langle S, A, \Phi, P, R \rangle$, where $S$ is a set of states, $A$ is a set of actions, $\Phi \subset S \times A$ is the set of admissible state-action pairs, $P:\Phi \times S \to [0, 1]$ is the transition probability function, and $R: \Phi \to \mathbb{R}$ is the expected reward function where $\mathbb{R}$ is the set of real numbers. $P(s, a, s')$ denotes the transition probability from state $s$ to state $s'$ under action $a$, and $R(s, a)$ denotes the expected reward from taking action $a$ in state $s$.

The goal of learning is to maximize future rewards. Using a discount factor $\gamma \in (0, 1]$, the expected value of the discounted sum of future rewards at time $t$ is defined as 
$R_t = \sum_{k=0}^\infty \gamma^k r_{t+k}$, that is, the sum of discounted rewards from time $t$ onward. The action value function $Q^\pi(s,a) = \mathbb{E}[R_t|s_t=s, a]$ is the expected return after taking action $a$ in state $s$ and then following policy $\pi$, where $\pi$ maps $(s, a) \in \Phi$ to the probability of picking action $a$ in state $s$. The optimal action-value function $Q^*(s, a) = \max_\pi Q^\pi (s, a)$.

The action value function $Q$ can be represented by a function approximator. Herein, we use deep Q-learning (DQN) to approximate $Q^*$ with a neural network $Q(s, a; \theta)$, where $\theta$ are parameters of the neural network \cite{mnih2013playing, mnih2015human}. DQN has seen broad success and is the basis for many deep RL variants \cite{gu2016continuous, van2016deep, wang2016dueling}. 

The parameters are learned iteratively by minimizing a sequence of loss functions $L_i(\theta_i)$,
$$L_i(\theta_i) = \mathbb{E}(r + \gamma \max_{a'}Q(s', a';\theta_{i-1})-Q(s, a;\theta_i))^2.$$
This specific formulation is termed one-step Q-learning, because $s'$ is the state that succeeds $s$, but it can be relaxed to $n$-step Q-learning by considering rewards over a sequence of $n$ steps. Alternative solution methods to DQN include proximal policy optimization \cite{schulman2017proximal} and asynchronous advantage actor-critic A3C \cite{mnih2016asynchronous}, both of which learn the policy $\pi$ directly. 

\begin{table*}[t]
\centering
\ra{1.3}
\begin{tabular}{@{}ll@{}}
\toprule
Paper & Network Description(s) \\
\midrule
Ghanem and Chen \cite{ghanem2018reinforcement} & \emph{100 machine} local area network \\
Schwartz and Kurniawati \cite{schwartz2019autonomous} & \emph{50 machines} with unknown services and 18 machines with 50 services \\
Ghanem and Chen \cite{ghanem2020reinforcement} & \emph{100 machine} local area network \\
Chaudhary \textit{et al.} \cite{chaudhary2020automated} & Not reported \\
Yousefi \textit{et al.} \cite{yousefi2018reinforcement} & Attack graph with \emph{44 vertices and 43 edges} \\
Chowdary \textit{et al.} \cite{chowdary2020autonomous} & Attack graph with 109 vertices, edges unknown, and a \emph{300 host} flat network \\
Hu \textit{et al.} \cite{hu2020automated} & Attack graph with \emph{44 vertices and 52 edges} \\
\textbf{Our network} & Attack graph with \textbf{955 vertices and 2350 edges} \\
\bottomrule
\\
\end{tabular}
\caption{Network sizes in the literature.}
\label{table:size}
\end{table*}
\subsection{Penetration Testing}

Penetration testing is defined by Denis \textit{et al.} as, ``a simulation of an attack to verify the security of a system or environment to be analyzed . . . through physical means utilizing hardware, or through social engineering \cite{denis2016penetration}.'' They continue by emphasizing that penetration testing is not the same as port scanning. Specifically, if port scanning is looking through binoculars at a house to identify entry points, penetration testing is having someone actually break into the house.

Penetration testing is part of broader vulnerability detection and analysis, which typically combines penetration testing with static analysis \cite{bacudio2011overview, shah2015overview, chess2004static}. Penetration testing models have historically taken the form of either the flaw hypothesis model \cite{pfleeger1989methodology, weissman1995penetration}, the attack tree model \cite{salter1998toward, schneier1999attack}, or the attack graph model \cite{mcdermott2001attack, duan2008easy, polad2017attack}.

The flaw hypothesis model describes the general process of gathering information about a system, determining a list of hypothetical flaws, e.g., via domain expert brain-storming, sorting that list by priority, testing hypothesized flaws in order, and fixing those that are discovered. As McDermott notes, this model is general enough to describe almost all penetration testing \cite{mcdermott2001attack}. The attack tree model adds a tree structure to the process of gathering information, generating hypotheses, etc., which allows for a standardization of manual penetration testing, and also gives a basis for automated penetration testing methods. The attack graph model adds a network structure, differing from the attack tree model in regard to the richness of the topology and, accordingly, the amount of information needed to specify the model.

Automated penetration testing has become a part of practice \cite{stefinko2016manual}, with the attack tree and attack graph models as its basis. In reinforcement learning, these models serve as the environment $\mathcal{E}$. They are depicted in Figure \ref{fig:1}C and \ref{fig:1}D, respectively. Both modeling approaches involve constructing topologies of networks by treating machines (i.e., servers and network devices) as vertices and links between machines as edges between vertices. Variants involve integrating additional detail regarding sub-networks and services. In the case of attack trees, probabilities must be assigned to the branches between parent and child nodes, and in the case of attack graphs, transition probabilities between states must be assigned to each edge. While many of the favorable properties of attack trees persist in attack graphs, it is unclear whether attack graphs can outperform attack trees in largely undocumented systems, i.e., systems with partial observability \cite{mcdermott2001attack, shmaryahu2016constructing}.

\subsection{Reinforcement Learning for Penetration Testing}

Reinforcement learning in penetration testing is promising because it addresses many challenges. A single penetration testing tool has never been enough \cite{austin2011one}. Yet, RL can be the basis for many tools, such as analysis, bypassing security, and penetration, and can by applied to the various types of penetration testing, i.e., external testing, internal testing, blind testing, and double-blind testing \cite{weissman1995security}. The automation and generality of RL means it can be deployed quickly, in the form of many variants with different policies, at many points in a network. And, as Chen \textit{et al.} note \cite{chen2018penetration}, as future networks scale in the Internet of Things age, intelligent payload mutation and intelligent entry-point crawling, the kinds of tasks RL is well-suited for, will be necessary in penetration testing.

Reinforcement learning for penetration testing uses the attack graph model \cite{ghanem2018reinforcement, schwartz2019autonomous, ghanem2020reinforcement, chaudhary2020automated, yousefi2018reinforcement, chowdary2020autonomous, hu2020automated}. The environment $\mathcal{E}$ is treated as either a MDP, mirroring classical planning, where actions are deterministic and the network structure and configuration are known, or as a Partially Observable Markov Decision Process (POMDP), where the outcomes of actions are stochastic and network structure and configuration are uncertain.

While POMDPs are more realistic, they have not been shown to scale to large networks and require modeling many prior probability distributions \cite{shmaryahu2016constructing}. Since full observability leads MDPs to underestimate attack cost, its main flaw is in finding vulnerabilities which are unlikely to be found or exploited. As such, penetration testing on MDPs gives a worst case analysis, making it the risk averse option in the sense that it tends towards false alarms. We use MDPs for our attack graph model because it can scale and because our methodology for adding cyber terrain to MDP attack graphs can later be extended to POMDPs.

Unlike most previous work in RL for penetration testing \cite{ghanem2018reinforcement, schwartz2019autonomous, ghanem2020reinforcement, chaudhary2020automated}, but similar to Yousefi et al., Hu et al., and Chowdary et al. \cite{yousefi2018reinforcement, hu2020automated, chowdary2020autonomous}, we use vulnerability information to construct the MDP. In particular, we  use the Common Vulnerability Scoring System (CVSS) \cite{gallon2011using}. Unlike those previous authors, however, we extend beyond vulnerability information by folding in notions of cyber terrain. Following the literature, we use DQN as the RL solution method \cite{schwartz2019autonomous, chowdary2020autonomous, hu2020automated}. However, we use a larger network than those in the literature, as reported in Table \ref{table:size}.

\section{Methods}

\begin{figure*}[t]
    \centering
    \includegraphics[width=\textwidth]{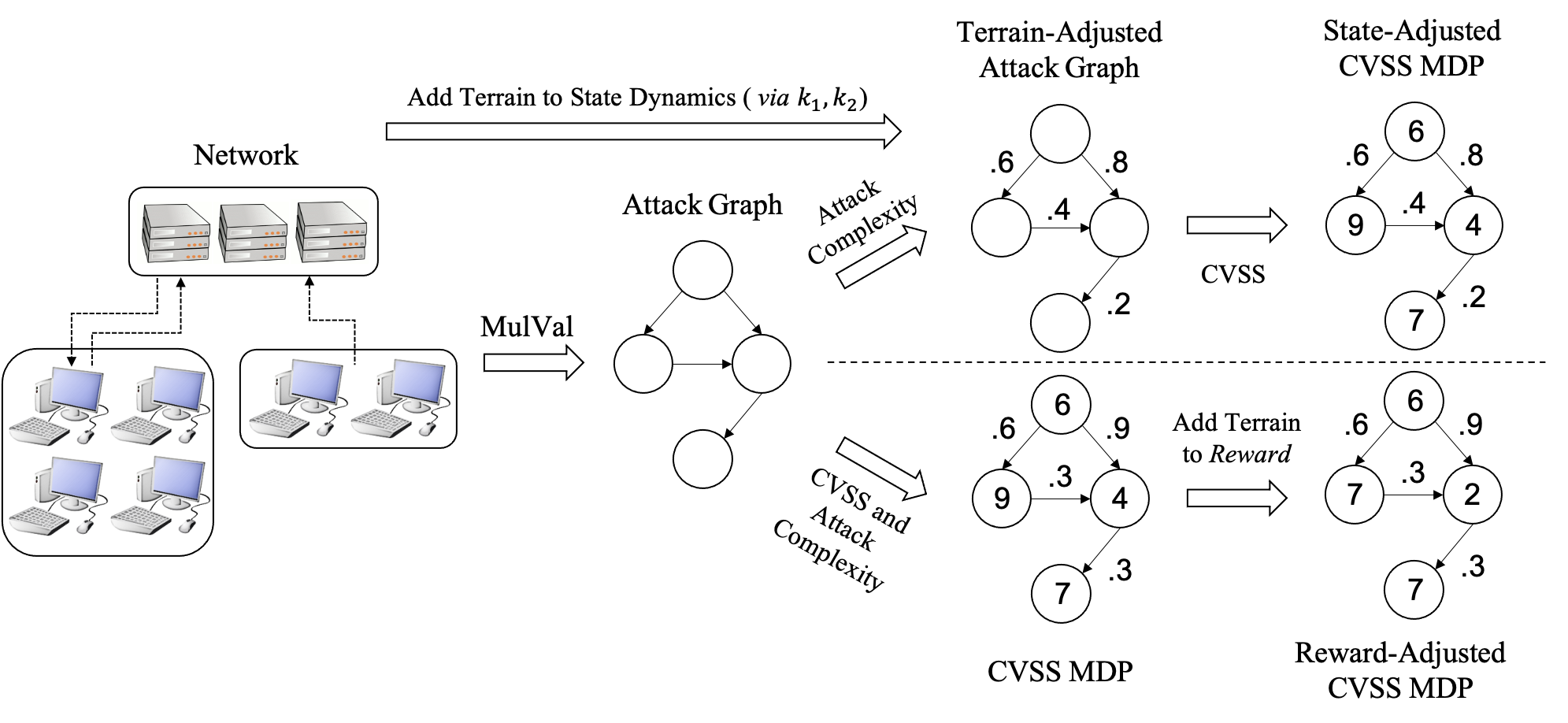}
    \caption{The network is extracted into an attack graph using MulVal \cite{ou2005mulval}. Terrain can be added via state or reward. To add via state, the attack graph is first modified to include more state information related to OAKOC \cite{conti_raymond_2018}. The CVSS MDP is constructed as usual with the transition terrain-adjusted probabilities. To add via reward, the CVSS MDP is constructed as usual followed by including terrain-adjusted rewards. Each method leads to a terrain-adjusted CVSS MDP. Note, attack complexity is a component of CVSS. The values inside the nodes correspond to reward values and the values on the edges are transition probabilities.}
    \label{fig:2}
\end{figure*}

RL-based penetration testing involves a three-step procedure of (1) extracting the network structure into an attack graph, (2) specifying an MDP (or POMDP) over the extracted attack graph, and (3) deploying RL on the MDP. The outcome of deploying RL can then be studied in various ways to express the penetration testing results.

The attack graph is extracted using MulVal, a framework that conducts multihost, multistage vulnerability analysis on a network representation using a reasoning engine \cite{ou2005mulval}. The states $S$ of the MDP are given by the vertices of the attack graph, which can be components of the network, e.g., entries into a specific subnet or an intermediary file server, or can be means of traversal, e.g., the interaction rules between network components. That is, not all states are locations in the network. The actions $A$ of the MDP that are available in a particular state are given by the outbound edges from that state.

The transition probabilities $P(s, a, s')$ and the reward $R$ of the MDP are constructed using CVSS. The transition probabilities are assigned using the attack complexity associated with $s'$, which CVSS ranks as either low, medium, or high, and which we translate into transition probabilities of $0.9$, $0.6$, and $0.3$, respectively, in following with Hu \textit{et al.} \cite{hu2020automated}. The agent remains in $s$ if the action fails. The reward for arriving at $s'$ is given by 
$$\emph{Base Score} + \frac{\emph{Exploitability Score}}{10}.$$
Then, a target node in the network is deemed the terminal state and given a reward of $100$. An initial state is defined and given a reward of $0.01$, and, using a depth first search, reward is linearly scaled from the initial state to the terminal state. Lastly, $-1$ reward is assigned to actions which bring the agent to a state from which the terminal state is inaccessible without backtracking, or otherwise lead to entering a sub-network from which the terminal state is not reachable.

We term this particular MDP the \emph{CVSS MDP}. The RL agent is trained using DQN in an episodic fashion. Episodes terminate when the terminal state is reached or after taking a number of hops, i.e., actions, in the network. This formulation is similar to those of Yousefi \textit{et al.}, Hu \textit{et al.}, and Chowdary \textit{et al.} \cite{yousefi2018reinforcement, hu2020automated, chowdary2020autonomous}. This terrain-blind approach ignores the typical perspectives of attackers when traversing and navigating enterprise networks.

Our methodology for adding cyber terrain builds on this formulation. We propose to add terrain via state and reward to resolve its short-comings in realism. To add terrain information via state is to do so by modifying $S$ and $P(s, a, s')$. First, additional information must be included from MulVal and other sources into the attack graph originally generated by MulVal. Then, this additional state information can be used to modify $S$ or $P(s, a, s')$. By using state, we represent terrain as an effect on the dynamics of the MDP. As such, it adds terrain by creating a more realistic model of the environment $\mathcal{E}$. To add terrain information via reward is to do so by modifying $R$, i.e., by reward engineering. Depending on the OAKOC phenomena, this means incrementing or decrementing the reward. By using reward, we introduce terrain not by directly bringing realism to $\mathcal{E}$, but rather by incentivizing the agent to behave in a more realistic manner. These two processes are depicted in Figure \ref{fig:2}. We term these \emph{terrain-adjusted} CVSS MDPs. The way this is done within our experiment is by assessing the network structure of the attack graph gathered from MulVAL initially. Then, firewalls, subnets and allowed/blocked services are parsed accordingly from the MulVAL output. Afterwards, this information is populated into a YAML file that captures, in hash map format, the relations between subnets and services that are allowed or blocked with firewalls. This YAML file is used to manipulate state and reward using scripts that adjust the rewards and transition probabilities based on the presence of a firewall and type of service blocked. This process of parsing out aspects of network structure then using them to create terrain-adjusted CVSS MDPs can be applied to terrain generally. The following subsections elaborate for this case of treating firewalls as obstacles.

The training process utilized a deep Q-network reinforcement learning agent using experience replay and a target Q-Network. This leveraged a flat action space, with  transformed layers with a ReLu activation function during a forward pass. An Adam optimizer was used, and smooth L1 loss was also used to ensure training stability. There were 850,000 training steps, among which 300,000 were exploration-based, with a replay buffer size of 50,000. The learning rate was 0.001, with a batch size of 32, and hidden layer sizes of 64 neurons. An initial epsilon of 1.0, with a final epsilon of 0.05 was also used to assist in training. Our next step will be to explain how the rewards or states were modified with the presence of firewalls, specifically based on the type of protocol.

\subsection{Firewalls as Obstacles}

We now consider firewall as a cyber terrain obstacles. Conti and Raymond categorize obstacles as physical or virtual capabilities that filter, disrupt, or block traffic between networks using different methods \cite{conti_raymond_2018}. For the purposes of this work, we consider firewalls as blocking obstacles and use the presented methodology to incorporate cyber terrain into a CVSS-based attack graph.

\subsubsection{Adding via Reward}

We engineer the reward to incentivize realistic attack campaigns using a term $k$ such that the reward in state $s$ after taking action $a$ becomes $$R(s, a) = R(s, a) + k(s).$$ The term $k$ decrements the reward to incentivize avoiding firewalls. The value of $k$ is dependent on the protocol, i.e.,
$$
k(s) =
\begin{cases}
                                 0 & \text{if no firewall} \\
                                 0.8w & \text{if FTP} \\
                                 0.6w & \text{if SMTP} \\
                                 0.4w & \text{if HTTP} \\
                                 0.2w & \text{if SSH} \\
\end{cases}
$$
where $w \leq 0$ is a parameter for tuning the strength of incentivization. That is, we vary the change in reward based on the security of the communication protocol. Note, when multiple protocols are blocked, their $k$ values are averaged together. The various $k$ values, dependent on protocol, are set in our paradigm using the criticality of each service to a firewall in a red-teaming exercise. For example, FTP is given a $k$ multiplier value of 0.8, whereas SSH is given a $k$ multiplier value of 0.2. Penetrating a host containing an FTP-based firewall along its path projections is likely to be harder than that of SSH, so we incentivize avoiding a firewall containing FTP more than that of SSH.

Consider the flow from the left to the bottom right of Figure \ref{fig:2}. To create the reward-adjusted MDP given an attack graph, first the CVSS MDP is constructed. Reward is assigned to the nodes using vulnerability scores and transition probabilities are assigned to the edges using exploitability scores as either 0.9, 0.6, or 0.3. This is depicted by the first step from the attack graph to the lower right steps in Figure \ref{fig:2}. Then, this CVSS MDP has its rewards modified corresponding to the presence of firewalls as depicted by the different reward values in the nodes in the final step of the lower right of Figure \ref{fig:2}, resulting in the reward-adjusted MDP.

\subsubsection{Adding via State}

Alternatively, we introduce realism by engineering the state transition probabilities. We use two terms $k_1(s)$ and $k_2(s)$ such that the state transition probabilities become $$P(s, a, s') = P(s, a, s') * k_1(s') * k_2(s').$$ The term $k_1$ corresponds to firewall presence and $k_2$ to the importance of the firewall. They are defined as follows.
$$
k_1(s) = \begin{cases}
             0.01  & \text{if firewall} \\
             1.0  & \text{else} \\
       \end{cases} \quad
k_2(s) = \begin{cases}
            1.0 & \text{if no firewall} \\
            0.2 & \text{if FTP} \\
            0.4 & \text{if SMTP} \\
            0.6 & \text{if HTTP} \\
            0.8 & \text{if SSH} \\
       \end{cases}
$$
Recall, $P(s, a, s')$ is initialized using the low, medium, and high CVSS attack complexity classes. Note that, $k_1$ introduces an emphasis on avoiding firewalls and $k_2$ counterbalances that emphasis for high-value targets. Note, when multiple protocols are blocked, their $k_2$ values are averaged together.

Consider the flow from the left to the upper right of Figure \ref{fig:2}. To create the state-adjusted MDP given an attack graph, first the transition probabilities given by exploitability scores are modified corresponding to the presence of firewalls. This is depicted by the first step from the attack graph to the upper right steps in Figure \ref{fig:2} where the attack graph has transition probabilities on its edges but no reward. Then, reward is assigned using the vulnerability score, resulting in the state-adjusted MDP, shown in the final step of the upper right of Figure \ref{fig:2}.

\section{Results}

We now compare the performance of DQN across (1) the vanilla, terrain-blind CVSS MDP, (2) the reward-adjusted MDP, enhanced via $R$, and (3) the state-adjusted MDP, enhanced via state transition probabilities $P(s, a, s')$. We use a 122 host network whose attack graph has 955 vertices and 2350 edges. All presented results use $w=-2$. The top-line results are shown in Table \ref{table:multi}. The introduction of terrain increases the number of hops, as agents must now navigate around firewalls. 

We can compare the reward as well. Note the reward functions are identical between the vanilla MDP and the state-adjusted MDP, but are different between the vanilla MDP and reward-adjusted MDP. The $w$ parameter for adjusting $R$ decreases reward, and so we expect to see a lower reward. Notably, we see a decrease in reward despite taking almost 30 more hops. Again, this simply confirms the reward has been decremented. 

\begin{table}[t]
\centering
\ra{1.3}
\begin{tabular}{@{}lccc@{}}
\toprule
MDP & ``Vanilla" & via $R$ & via $P(s, a, s')$ \\
\midrule
Total Number of Hops & 62 & 91 & 85 \\
Total Reward & 221 & 179 & 237 \\
\bottomrule
\\
\end{tabular}
\caption{Total number of hops and total reward with all protocols available.}
\label{table:multi}
\end{table}

Similar comparisons between the vanilla MDP and state-adjusted MDP, we see a greater reward, as expected due to the larger number of hops. Whereas the agent averages ~3.6 units of reward per hop on the vanilla MDP, the agent averages ~2.8 units of reward per hop on the state-adjusted MDP. Recalling that reward for approaching the terminal state is linearly scaled using a depth first search from the initial state to terminal state, the maintenance of a high average reward suggests the RL agent can still make steady progress to the terminal state while accounting for obstacles.

\begin{figure*}[t]
    \centering
    \includegraphics[width=\textwidth]{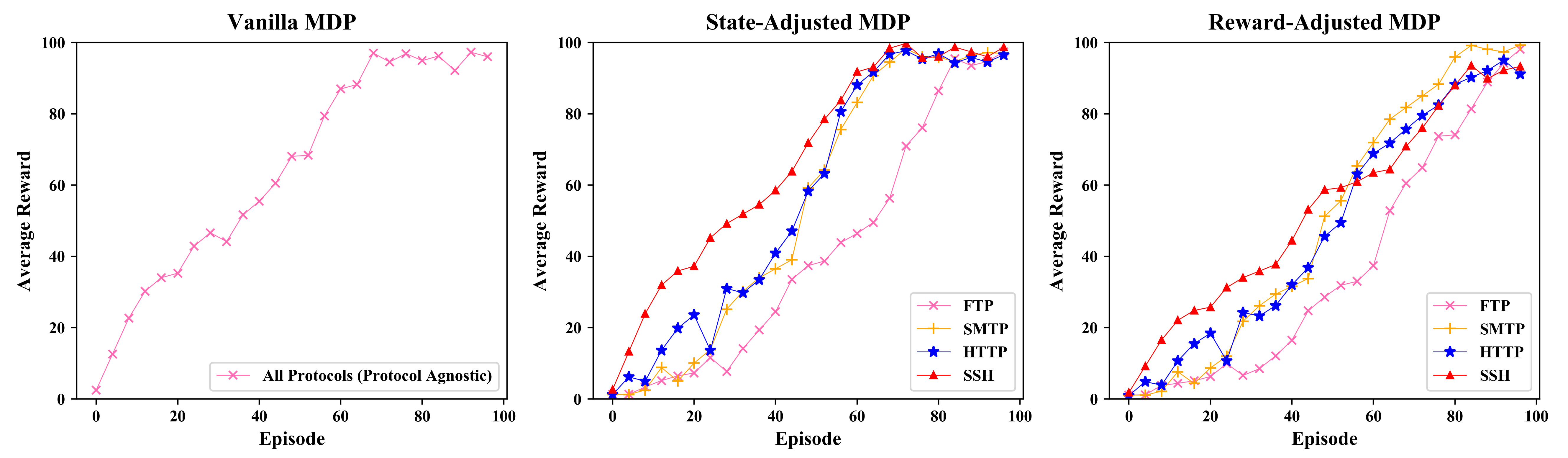}
    \caption{Average reward plotted against training episode for each MDP. The middle and right plots show special cases where the state-adjusted and reward-adjusted MDPs were restricted to a single communication protocol.}
    \label{fig:plots}
\end{figure*}

A closer look at the results is shown in Figure \ref{fig:plots}. The plots show the average reward achieved by the DQN agent against the number of training episodes. The total reward was evaluated every 4 episodes and each episode had a maximum length of 2500 steps. The high average reward values achieved after 80 episodes signify that the agents spend a majority of their time close to the terminal state. The vanilla MDP is protocol agnostic. While Table \ref{table:multi} shows state-adjusted and reward-adjusted total reward when agents can choose between protocols, in Figure \ref{fig:plots}, the state-adjusted and reward-adjusted plots show average reward when the agent is restricted to a single choice of protocol. The plots show our method was able to represent that FTP is a more significant cyber obstacle than SSH.

\begin{figure}[t]
    \centering
    \includegraphics[width=5cm, height=8cm]{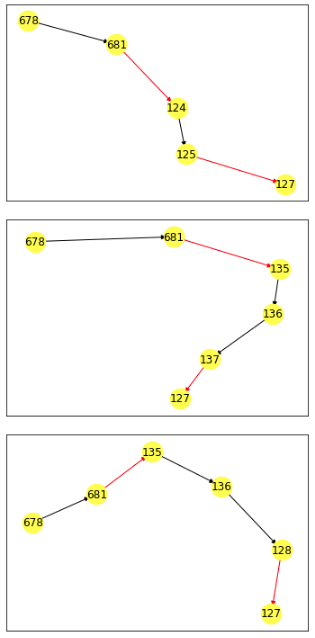}
    \caption{Visualization of attack campaigns.}
    \label{fig:network_paths}
\end{figure}

Lastly, Figure \ref{fig:network_paths} shows the paths derived from the approximated policies. The top figure shows the vanilla path, the middle figure show the state-adjusted path, and the bottom figure shows the reward-adjusted paths. The paths have been greatly reduced by focusing on key nodes along the path. The red edges highlight differences in the path taken from the initial to terminal state. At node 681, a firewall existed that led to the agents using state-adjusted and reward-adjusted MDPs to seek an alternate path. Their paths differentiate after node 136.

\section{Remarks}

The presented reward- and state-adjusted MDP construction methods build on the existing practice of constructing MDPs using the CVSS. Importantly, the presented methodology maintains the scalability of CVSS MDPs while folding in richer aspects of network and path structures. Historically, constructing attack graphs was largely a manual process \cite{mcdermott2001attack}. Cyber terrain was included by operators manually in these detailed, tedious, and unscalable MDPs. This tradition has continued in the construction of MDPs for RL for penetration testing \cite{ghanem2018reinforcement, schwartz2019autonomous, ghanem2020reinforcement, chaudhary2020automated}. The use of MulVal to construct attack graphs marked a broader change in philosophy, however, where scalability of attack graph construction was valued over realism \cite{ou2005mulval, gallon2011using}. This has been mirrored by the use of the CVSS to scale MDP construction alongside attack graph construction \cite{yousefi2018reinforcement, hu2020automated, chowdary2020autonomous}. The use of CVSS to scale MDP construction, like the broader shift in philosophy marked by MulVal, sacrifices realism. The presented methodology offers scalable approaches for building realism into scalable approaches to the construction of MDPs over attack graphs by drawing on notions of cyber terrain from intelligence preperation of the battlefield.

\section{Conclusion}

In this paper, we present methods for enhancing ``vanilla", CVSS-based attack graphs by using concepts of cyber terrain within intelligence preparation of the battlefield. Our method introduces cyber terrain by modifying the state transition probabilities $P(s, a, s')$ and reward function $R$. Using an example attack graph with nearly 1000 nodes, we showed how our approach can be used to introduce cyber obstacles, particularly firewalls. We evaluated using DQN, and showed notable differences in total reward, number of hops, average reward, and attack campaigns.

The shift from manually constructed MDPs \cite{ghanem2018reinforcement, schwartz2019autonomous, ghanem2020reinforcement, chaudhary2020automated} to CVSS-based MDPs \cite{yousefi2018reinforcement, chowdary2020autonomous, hu2020automated} marks an emphasis on scaling the construction of attack-graph-based MDPs. Our methodology maintains an automated, scale-oriented approach to constructing MDPs, while introducing notions of cyber terrain that help ground RL agent behavior to reality.

Future work should consider how more elements of cyber terrain can be folded into MDP construction. In doing so, a primary consideration should be to continue to scale the size of attack graphs, using more hosts at an enterprise scale. This would help further validate the use of cyber-terrain IPB principles in creating realistic contexts for penetration testing. Also, methods should be developed that use multiple initial and terminal states to assist in attack surface cartography. In addition, the current literature considers RL agents that are trained and deployed on the same network. Notions of transfer learning, meta-learning, and lifelong learning are promising paths for generalizing penetration testing agents.



\bibliographystyle{IEEEtran}
\bibliography{ref}

\end{document}